\documentclass[letterpaper, 10 pt, conference]{ieeeconf}  

\IEEEoverridecommandlockouts                              

\usepackage{graphics} 
\usepackage{graphicx}
\usepackage{epsfig} 
\usepackage{mathptmx} 
\usepackage{times} 
\usepackage{amsmath} 
\usepackage{amssymb}  
\usepackage{algorithm}
\usepackage{algpseudocode}
\usepackage[capitalize]{cleveref} 
\usepackage{subcaption}
\usepackage{booktabs}
\usepackage{array}
\usepackage{tikz}
\usetikzlibrary{positioning}
\usepackage{multirow}

\usepackage{amsthm}

\newtheorem{theorem}{Theorem}

\theoremstyle{remark}
\newtheorem{remark}{Remark}

\title{\LARGE \bf
Environment Design for Reliable Shared Autonomy \\with Probabilistic Guarantees}

\begin{document}

\author{Yi-Shiuan Tung, Himanshu Gupta, Gyanig Kumar, Heyang Huang, Bradley Hayes, Alessandro Roncone\\
Department of Computer Science, University of Colorado Boulder\\ \{\tt\small yi-shiuan.tung, himanshu.gupta, gyanig.kumar, heyang.huang, bradley.hayes, alessandro.roncone\}\\@colorado.edu}


\maketitle
\thispagestyle{empty}
\pagestyle{empty}

\begin{abstract}
Shared autonomy enables humans and robots to collaboratively perform tasks by combining human input with autonomous assistance. Most prior work focuses on improving intent inference under a fixed environment, overlooking how workspace design itself affects inference difficulty. We observe that the physical arrangement of objects directly influences the separability of candidate goals under noisy user inputs. We formulate workspace design as an optimization problem and derive a probabilistic correctness guarantee under a bounded noise model. Through simulation experiments across multiple tabletop scenarios, we show that optimized layouts improve goal inference reliability and reduce ambiguity compared to baseline arrangements. We further demonstrate a real-world shared autonomy system that integrates the proposed inference framework. This highlights the role of environment design as a complementary axis for improving shared autonomy systems.
\end{abstract}

\section{Introduction \& Related Works}
Shared autonomy enables humans and robots to collaboratively control a robotic system by blending human input with autonomous assistance. Human control remains crucial in unstructured or safety-critical environments, where full autonomy is unreliable due to factors such as perceptual uncertainty, task ambiguity, or the need to leverage human expertise, as commonly seen in assistive robotics \cite{eirale2025human, fu2025tasc} and surgical robotics \cite{ballantyne2003vinci}. However, pure teleoperation can impose a high cognitive and physical burden on the operator, since a low-dimensional interface (e.g., a joystick or keyboard) must be used to control a high-dimensional robotic system, such as a 7-DoF manipulator. To alleviate this burden, shared autonomy systems aim to infer the user’s intent from observed inputs and assist in task execution once sufficient confidence is achieved~\cite{liu2025casper}. Prior work has focused on improving intent inference through algorithmic advances, including planning under uncertainty~\cite{javdani2018shared} and leveraging vision-language models (VLMs) to identify user goals~\cite{liu2025casper}. However, these approaches treat the environment as fixed and focus on improving inference given a particular scene.
\begin{figure}[t]
    \centering
    \begin{subfigure}[b]{0.45\textwidth}
        \centering
        \includegraphics[width=\textwidth]{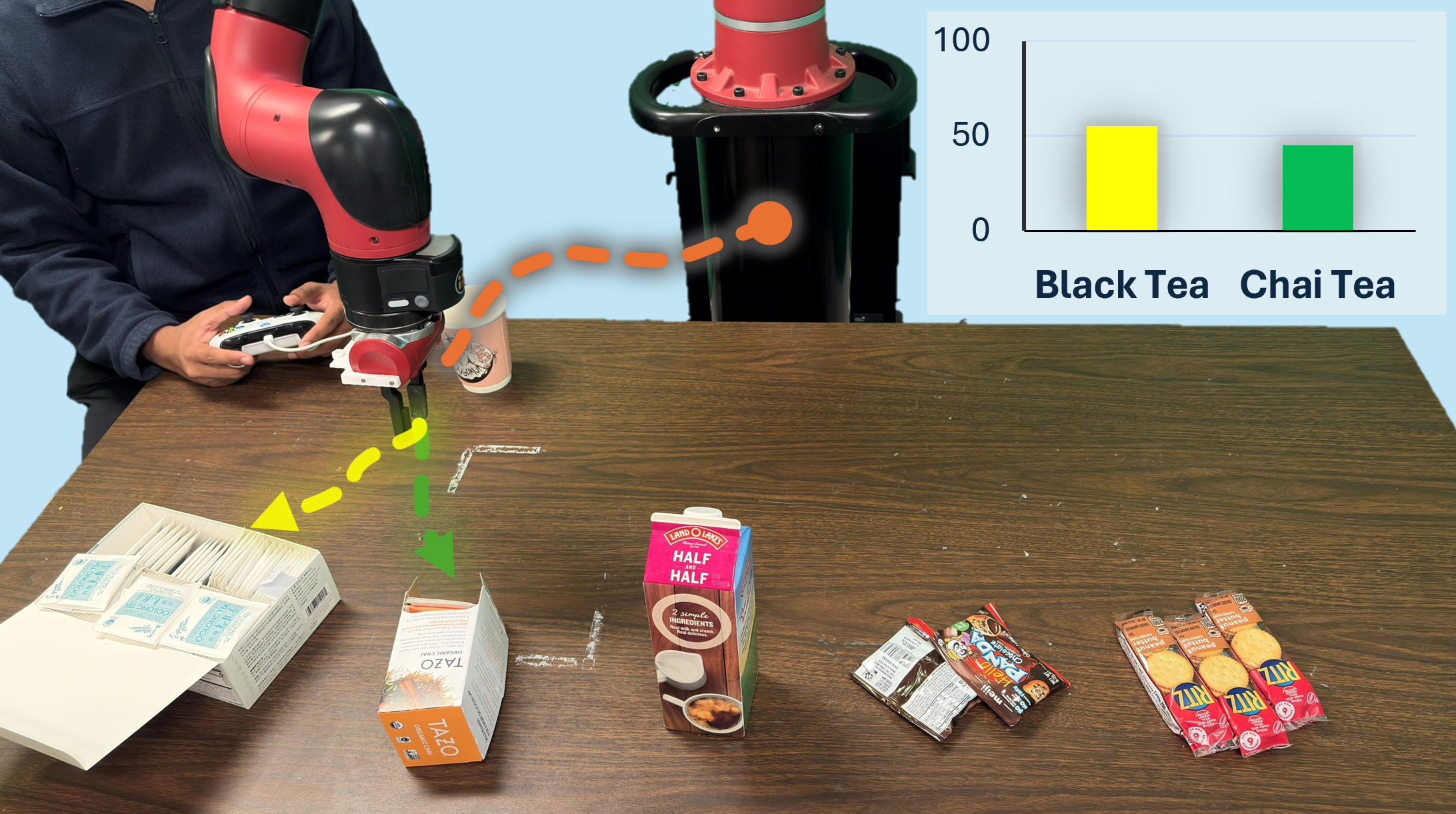}
        \caption{\textbf{Baseline setup.} Without workspace optimization, the robot struggles to accurately infer which tea bag the user intends to pick up, leading to ambiguous goal predictions.}
        \label{fig:setup1}
    \end{subfigure}
    \vskip\baselineskip
    \begin{subfigure}[b]{0.45\textwidth}
        \centering
        \includegraphics[width=\textwidth]{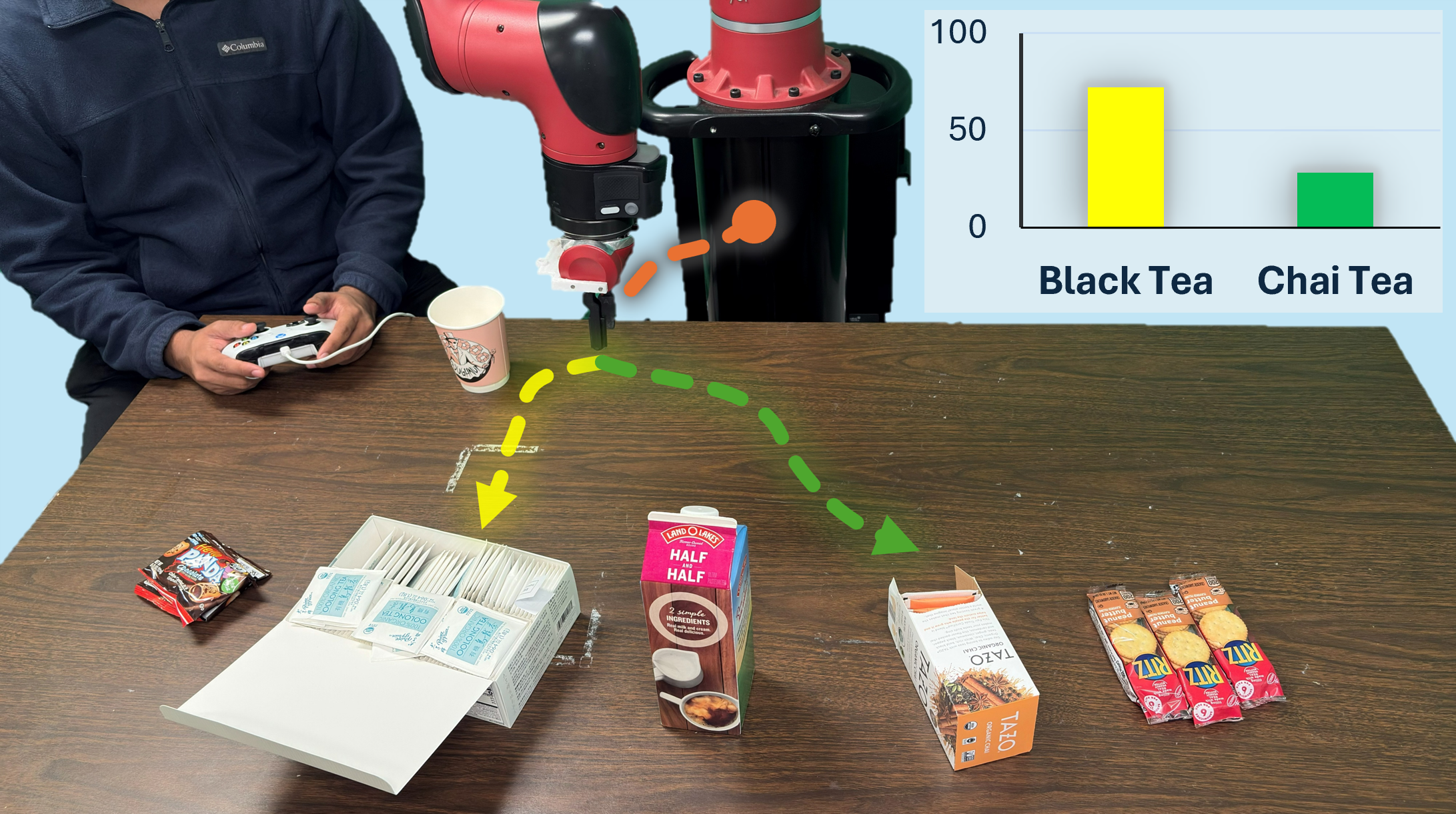}
        \caption{\textbf{Optimized workspace.} The robot quickly infers that the user is picking up a black tea bag to place in the cup.}
        \label{fig:setup3}
    \end{subfigure}
    \caption{Given an initial trajectory (orange), the robot infers a probability distribution over possible goals (yellow and green). Our proposed workspace optimization rearranges objects to enhance the accuracy and speed of intent prediction.}
    \label{fig:intro}
\end{figure}

Our key insight is that the physical arrangement of objects in the workspace shapes the difficulty of the inference problem. Object placement determines how distinguishable different goals are under noisy human inputs, and therefore directly affects both the speed and reliability of intent inference. This motivates our complementary approach: instead of only improving inference algorithms, we optimize the workspace to make user intent more legible. Prior work formulates workspace optimization by maximizing the legibility of candidate goals along a nominal trajectory, encouraging object placements that reveal the user’s intent early \cite{tung2024workspace, Tung2023HRI}. This is achieved by minimizing the cross-entropy between the inferred goal distribution and the true goal over time. However, this formulation assumes a deterministic nominal trajectory and does not explicitly model uncertainty in user inputs. As a result, the optimization does not account for how execution noise perturbs the trajectory and affects goal distinguishability. In contrast, we seek to design workspaces that are robust to user input uncertainty.

In this work, we formulate workspace design as an optimization problem over object placements that maximizes the separability of candidate goals under a probabilistic user model. Building on a bounded Gaussian model of joystick noise, we derive an objective that provides a probabilistic correctness guarantee: the system identifies the correct goal with probability at least $1 - \alpha$, where $\alpha$ is a user-specified failure probability. This yields a principled connection between environment design, inference uncertainty, and performance guarantees. We evaluate our method in simulation and demonstrate that optimized workspaces improve goal inference reliability compared to baselines. We further outline real-world tabletop scenarios for future user studies, where we will assess the impact of workspace design on task efficiency and user experience. This work makes the following contributions:
\begin{itemize}
  \item We formulate workspace design for shared autonomy as an optimization problem that explicitly accounts for intent inference under noise.
  \item We derive a probabilistic correctness guarantee linking workspace geometry to goal inference reliability.
  \item We demonstrate empirically that optimized workspaces improve inference performance in simulation.
\end{itemize}

\section{Workspace Optimization for Intent Inference} \label{sec:workspace_optimization}
We formalize the problem of arranging tabletop objects to enable a shared autonomy system to infer the human operator's intended goal both rapidly and reliably. The key insight is that
object placement is a controllable design variable: by selecting object configurations, we directly shape the geometry of the underlying inference problem. 

\subsection{Preliminaries}
\label{sec:preliminaries}

Consider a set of $M$ candidate goal objects at positions $\theta = \{\mathbf{g}_1, \dots, \mathbf{g}_M\} \subset \mathbb{R}^d$ on a shared workspace, and a robot end-effector (EE) initialized at $\mathbf{x}_0 \in \mathbb{R}^d$. At each discrete timestep $t$, the human provides a noisy control input
\begin{equation}
  \mathbf{u}_t = \boldsymbol{\pi}_h(\mathbf{x}_t, \mathbf{g}^*)
  + \boldsymbol{\epsilon}_t,
  \qquad
  \boldsymbol{\epsilon}_t \sim \mathcal{N}(\mathbf{0}, \boldsymbol{\Sigma}),
  \label{eq:observation_model}
\end{equation}
where $\mathbf{g}^* \in \theta$ denotes the user's intended goal, $\boldsymbol{\pi}_h(\mathbf{x}, \mathbf{g}) = (\mathbf{g} - \mathbf{x})/\|\mathbf{g} - \mathbf{x}\|$ is a unit vector pointing toward $g$, and $\boldsymbol{\Sigma}$ is the joystick noise covariance. Under first-order integrator dynamics, the EE evolves as $\mathbf{x}_{t+1} = \mathbf{x}_t + \mathbf{u}_t\,\Delta t$, producing a trajectory $\xi = \mathbf{x}_{0:T}$.

The robot maintains a belief over goals using the Boltzmann-rational model \cite{dragan2013legibility}, which assigns higher probability to goals that render the observed trajectory efficient according to a cost function $C$:
\begin{align}
  P(\mathbf{g} \mid \xi_{0:t})&
  \propto P(\mathbf{g})\,
  \exp\!\bigl(-\beta\, C(\xi_{0:t},\,\mathbf{g})\bigr)
  \label{eq:path_efficiency}
\end{align}
where $\beta > 0$ is a rationality coefficient. The system commits to a goal once the posterior confidence exceeds a threshold, $\max_g P(\mathbf{g} \mid \xi_{0:t}) \ge p_{\mathrm{thresh}}$. Following \cite{dragan2013legibility}, we define the cost function $C(\xi_{0:t}, g)$ as the sum of the trajectory length (under the $l_2$ norm) and the remaining distance-to-go to the goal.

\subsection{Workspace Optimization with Probabilistic Guarantees}
%
%
%
\label{sec:trajectory_margin}

To design workspaces that are robust to execution uncertainty, we propose a formulation that directly enforces \emph{probabilistic correctness} of goal inference under input noise. Our approach is based on a margin criterion that measures separation between competing goal hypotheses.

\paragraph{Cost difference and margin}
Consider the nominal trajectory $\xi_{0:T}$ toward the true goal $\mathbf{g}^*$. We define the pairwise cost difference between the true goal and an alternative $\mathbf{g}$ as
\begin{equation}
\delta c(\mathbf{g})
= C(\xi_{0:T}, \mathbf{g}) - C(\xi_{0:T}, \mathbf{g}^*)
\label{eq:cost_diff}
\end{equation}
which measures how much less efficient the trajectory is under $\mathbf{g}$ compared to $\mathbf{g}^*$. When $\delta c(\mathbf{g}) > 0$, the trajectory is more consistent with the true goal. 
Under the Boltzmann model, we define the log posterior margin ($M_g(\xi)$) as the log ratio of posterior probabilities between the true goal $\mathbf{g}^*$ and an alternative $\mathbf{g}$.
\begin{equation}
  M_g(\xi)
  =
  \log \frac{P(\mathbf{g}^* \mid \xi)}{P(\mathbf{g} \mid \xi)}
  =
  \beta\,\delta c(\mathbf{g})
  + \log \frac{P(\mathbf{g}^*)}{P(\mathbf{g})}.
\end{equation}
We assume uniform priors, which reduces $M_g$ to $\beta\,\delta c(\mathbf{g})$.

\paragraph{From margin to correctness}
Correct goal inference requires that the true goal has higher posterior probability than all alternatives:
\begin{equation}
M_g(\xi) > 0, \quad \forall \mathbf{g} \neq \mathbf{g}^*.
\end{equation}

In practice, execution noise makes the margin $M_g(\xi)$ a random variable. Therefore, we seek to enforce that this condition holds with high probability:
\begin{equation}
\mathbb{P}\!\left(M_g(\xi) > 0,\; \forall \mathbf{g} \neq \mathbf{g}^*\right)
\ge 1 - \alpha.
\end{equation}

\paragraph{Uncertainty from trajectory noise.}
In practice, joystick noise perturbs the executed trajectory, introducing uncertainty into the cost difference for goal inference. Under integrator dynamics with isotropic Gaussian control noise, the terminal state $x_T$ satisfies
\begin{equation}
  \mathbf{x}_T \sim \mathcal{N}(\bar{\mathbf{x}}_T, \Sigma_{x_T}),
  \qquad
  \Sigma_{x_T} = T\,\sigma_u^2\,\Delta t^2\,\mathbf{I},
  \label{eq:position_variance}
\end{equation}
where $\bar{\mathbf{x}}_T$ is the nominal terminal state obtained in the absence of noise, $\sigma_u^2$ is the variance of the joystick noise at each timestep, $T$ is the horizon length, $\Delta t$ is the timestep, and $\mathbf{I}$ is the identity matrix. The covariance $\Sigma_{x_T}$ captures how uncertainty accumulates over time. To understand how this uncertainty affects inference, we analyze the pairwise cost difference $\delta c(\mathbf{g})$ around the nominal endpoint $\bar{\mathbf{x}}_T$. Because $C(\xi, \mathbf{g})$ is generally nonlinear in the trajectory, $\delta c$ is not Gaussian even if $\mathbf{x}_T$ is Gaussian. To obtain a tractable approximation, we linearize $\delta c(\mathbf{g})$ with respect to the terminal state around the nominal endpoint:
\begin{equation}
  \delta c(\mathbf{g}; \mathbf{x}_T)
  \approx
  \delta c(\mathbf{g}; \bar{\mathbf{x}}_T)
  +
  \nabla_{\mathbf{x}_T} \delta c(\mathbf{g})\big|_{\bar{\mathbf{x}}_T}^{\!\top}
  (\mathbf{x}_T - \bar{\mathbf{x}}_T).
  \label{eq:delta_c_linearization}
\end{equation}
This approximation is locally valid when execution noise remains within a neighborhood of the nominal trajectory where higher-order terms are bounded; we formalize this condition and bound the linearization error in Appendix~\ref{app:linearization_validity}. Let $\mathbf{a}_g := \nabla_{\mathbf{x}_T} \delta c(\mathbf{g})\big|_{\bar{\mathbf{x}}_T}$ denote the sensitivity of the cost difference with respect to the terminal state. Under this first-order approximation, the pairwise log-posterior margin becomes Gaussian with variance $v_g = \beta^2 \, \mathbf{a}_g^\top \Sigma_{x_T} \mathbf{a}_g$.
This variance term captures how sensitive the goal separation is to execution noise: if small perturbations in the terminal state induce large changes in $\delta c(\mathbf{g})$, the corresponding goals are more easily confused.

\paragraph{Probabilistic guarantee}
We now connect this margin formulation to goal inference correctness. Recall that for each competing goal $\mathbf{g} \neq \mathbf{g}^*$, the pairwise log-posterior margin is modeled as a Gaussian random variable under the linearized approximation $M_g \sim \mathcal{N}(m_g, v_g)$ where $m_g := \mathbb{E}[M_g]$ is the nominal margin and $v_g$ is the margin variance induced by joystick noise. Correct identification of the true goal $\mathbf{g}^*$ requires that all pairwise margins are nonnegative: $M_g \ge 0 \quad \forall \mathbf{g} \neq \mathbf{g}^*$.

For a fixed competing goal $\mathbf{g}$, we bound the probability of misclassification using a Gaussian tail bound:
\[
\mathbb{P}(M_g < 0)
=
\Phi\!\left(-\frac{m_g}{\sqrt{v_g}}\right),
\] where $\Phi$ is the standard normal cumulative distribution function. To enforce $\mathbb{P}(M_g < 0) \le \alpha_g$, it suffices that \[ m_g \ge \Phi^{-1}(1-\alpha_g)\,\sqrt{v_g}.\]

Applying a union bound over all $M-1$ alternative goals, we allocate $\alpha_g = \alpha / (M-1)$ to each pairwise comparison and define the \emph{trajectory margin slack}:
\begin{equation}
  s_g(\theta)
  :=
  m_g(\theta)
  -
  \Phi^{-1}(1-\alpha_g)\,\sqrt{v_g(\theta)}.
  \label{eq:traj_slack}
\end{equation}
A positive slack ensures robustness of the margin against execution noise. We consider the setting in which the user pursues a fixed true goal $\mathbf{g}^*$ over the inference horizon, the candidate goal set is known and finite, and inference is performed using the Boltzmann-rational observer defined above. Under Gaussian joystick noise and a local linearization of the pairwise margin, we obtain the following sufficient condition for correct inference.
%
%
\begin{theorem}[Sufficient condition for correct goal inference]
Fix a workspace parameter $\theta$ and true goal $\mathbf{g}^*$.
Suppose:
\begin{enumerate}
  \item The trajectory induces a Gaussian terminal state $\mathbf{x}_T \sim \mathcal{N}(\bar{\mathbf{x}}_T, \Sigma_{x_T})$.
  \item For each $\mathbf{g} \neq \mathbf{g}^*$, the pairwise log-posterior
  margin $M_g$ admits a first-order linear approximation and is Gaussian: $M_g \sim \mathcal{N}(m_g, v_g)$.
  \item The local linearization error is negligible within a
  $(1-\alpha)$-probability neighborhood of $\bar{\mathbf{x}}_T$.
\end{enumerate}

If $s_g(\theta) > 0
  \ \forall \mathbf{g} \neq \mathbf{g}^*$
then the probability of correct goal inference satisfies
\begin{equation}
  \mathbb{P}\bigl(\hat{\mathbf{g}} = \mathbf{g}^*\bigr)
  \ge 1 - \alpha.
  \label{eq:global_guarantee}
\end{equation}
\end{theorem}

\begin{remark}
The slack $s_g(\theta)$ measures how robustly the true goal is separated
from an alternative $\mathbf{g}$ under execution noise. The mean term
$m_g$ captures nominal distinguishability between goals, while the
variance term $v_g$ penalizes sensitivity to noise. Ensuring positive
slack for all competing goals guarantees that the true goal remains the
most likely hypothesis with probability at least $1-\alpha$.
\end{remark}

\paragraph{Workspace design objective}
We optimize the workspace to maximize the worst-case margin slack across all possible true goals and competing alternatives:
\begin{equation}
  \max_{\theta}\;
  \min_{\mathbf{g}^*}\;
  \min_{\mathbf{g} \neq \mathbf{g}^*}\;
  s_g(\theta).
  \label{eq:design_objective}
\end{equation}
When this objective is positive, the system guaranties a correct identification of the goal with probability at least $1-\alpha$ for \emph{every}
goal, providing a design-time certificate of reliability of the inference.
\subsection{Quality-Diversity Optimization}
\label{sec:qd}

The workspace design objective~\eqref{eq:design_objective} is non-convex and multi-modal, with multiple disjoint regions of the layout space yielding comparable objective values. In such settings, standard optimization methods are prone to mode collapse or local maxima, returning a single suboptimal solution that does not reflect the full structure of the solution space. We use MAP-Elites \cite{mouret2015illuminating} which maintains an archive of solutions indexed by low-dimensional behavioral descriptors. Each cell in the archive stores the highest-quality solution observed for a given region of the behavior space. In our setting, the quality of a layout is measured by its worst-case trajectory margin slack, while the behavior descriptors capture geometric properties of the layout.

\paragraph{Behavior space}
To encourage diversity in spatial configurations, we define behavior space based on
\begin{enumerate}
  \item the mean pairwise distance between objects, which captures overall spatial spread, and
  \item the offset of the object centroid from the workspace center,
  which captures global asymmetry.
\end{enumerate}

\paragraph{Optimization}
We instantiate the QD framework using CMA-ME~\cite{fontaine2020covariance}, which leverages covariance matrix adaptation to efficiently explore the continuous layout space. The algorithm iteratively proposes candidate layouts, evaluates their margin slack, and inserts them into the archive based on their behavioral descriptors.

\paragraph{Constraints}
All candidate layouts are required to satisfy geometric feasibility constraints, including workspace bounds, inter-object clearance, and minimum distance from the robot's initial pose. These constraints ensure that archived solutions are physically realizable and safe for execution. Additional implementation details are provided in Appendix~\ref{app:qd}.
\begin{table}[t]
\centering
\caption{Evaluation environments. $M$ is the number of pick objects. ``Total'' includes fixed objects that are not picked but contribute clutter.}
\label{tab:envs}
\small
\begin{tabular}{@{}llccp{3cm}@{}}
\toprule
\textbf{Tier} & \textbf{Scene} & $M$ & \textbf{Total} & \textbf{Pick objects} \\
\midrule
Easy   & breakfast\_easy  & 2 & 3  & cereal box, yellow banana \\
Med-A  & desk             & 3 & 5  & white mug, black stapler, maroon pen cup \\
Med-B  & breakfast        & 3 & 6  & cereal box, yellow banana, milk carton \\
Med-C  & kitchen\_prep    & 3 & 5  & red apple, red can, green bottle \\
Hard-A & meal\_assembly   & 5 & 7  & cereal box, yellow banana, red apple, red can, green bottle \\
Hard-B & cluttered        & 8 & 11 & 8 colored blocks and cylinders \\
\bottomrule
\end{tabular}
\end{table}
\begin{figure*}[t]
\centering
\newcommand{\sceneimg}[1]{\includegraphics[width=0.22\linewidth]{figures/sim/#1}}
\begin{tabular}{cccc}
\textbf{Baseline} & \textbf{Optimized} &
\textbf{Baseline} & \textbf{Optimized} \\
\sceneimg{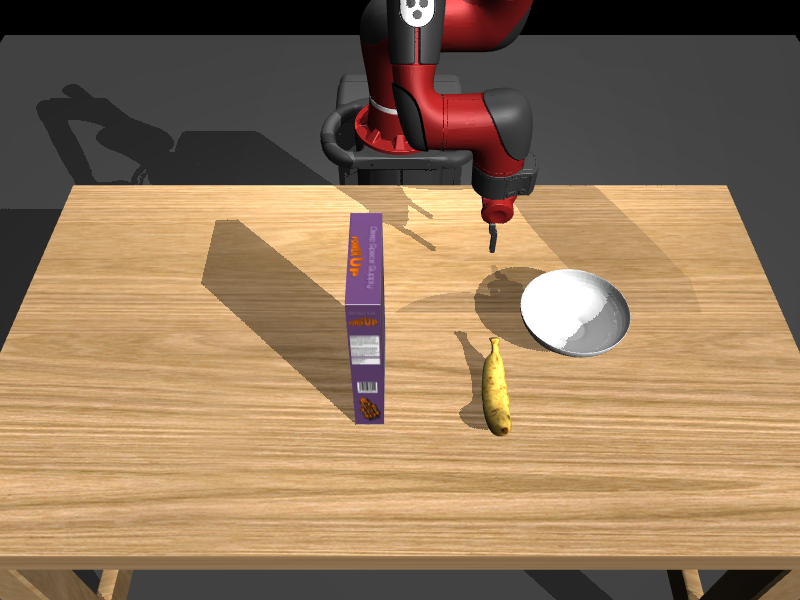} &
\sceneimg{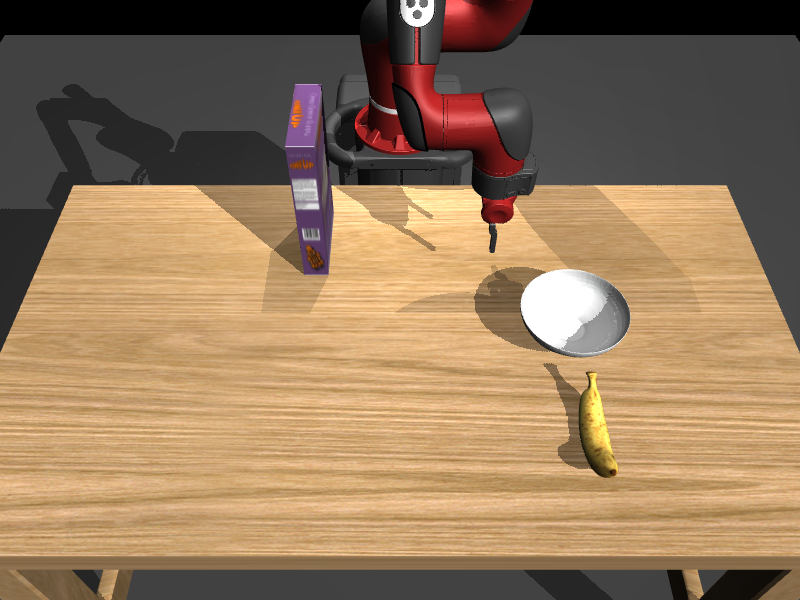} &
\sceneimg{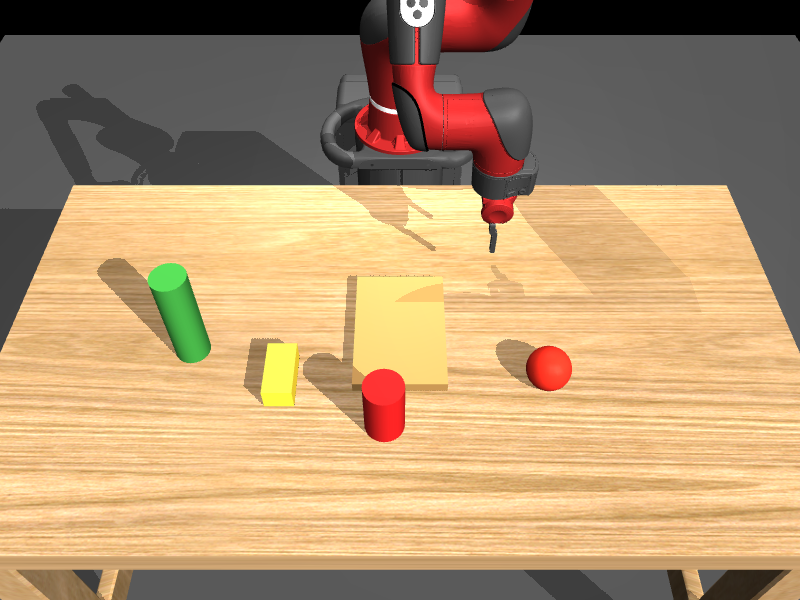} &
\sceneimg{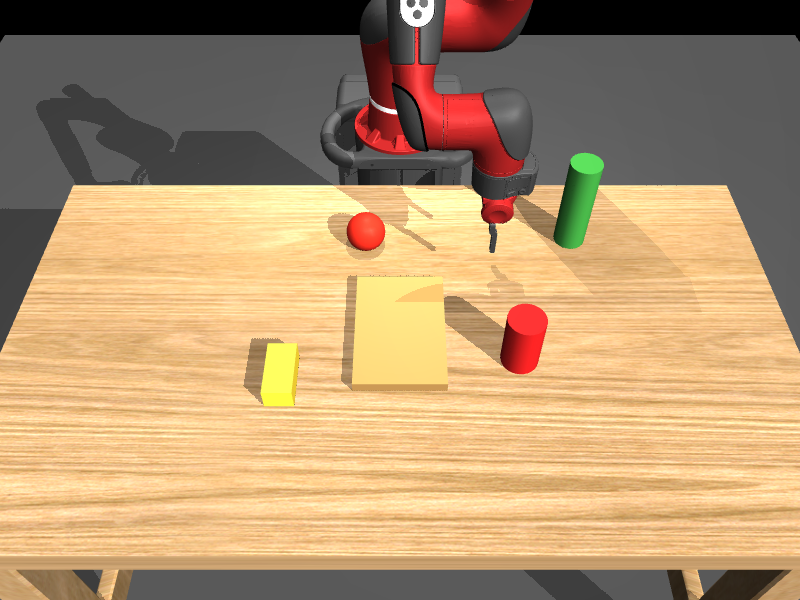} \\
\multicolumn{2}{c}{(a) Easy: \emph{breakfast\_easy} ($M{=}2$)} &
\multicolumn{2}{c}{(d) Med-C: \emph{kitchen\_prep} ($M{=}3$)} \\[0.6em]
\sceneimg{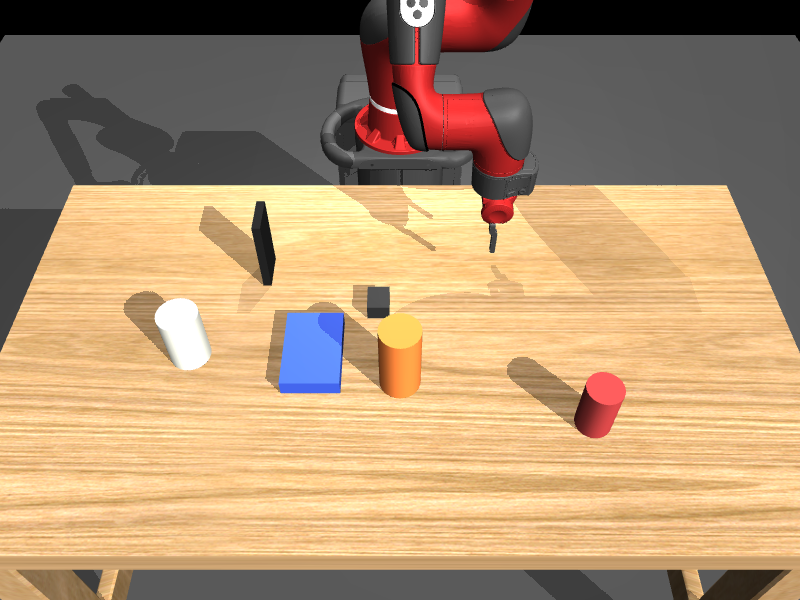} &
\sceneimg{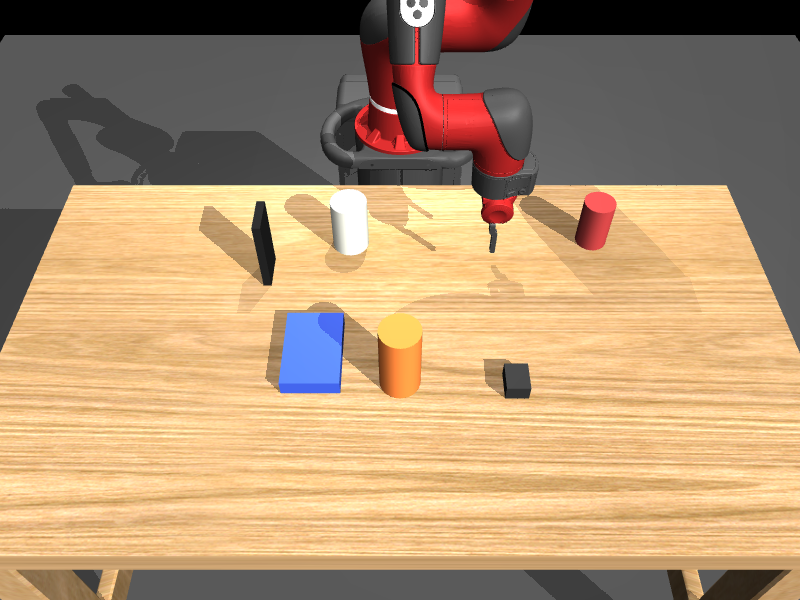} &
\sceneimg{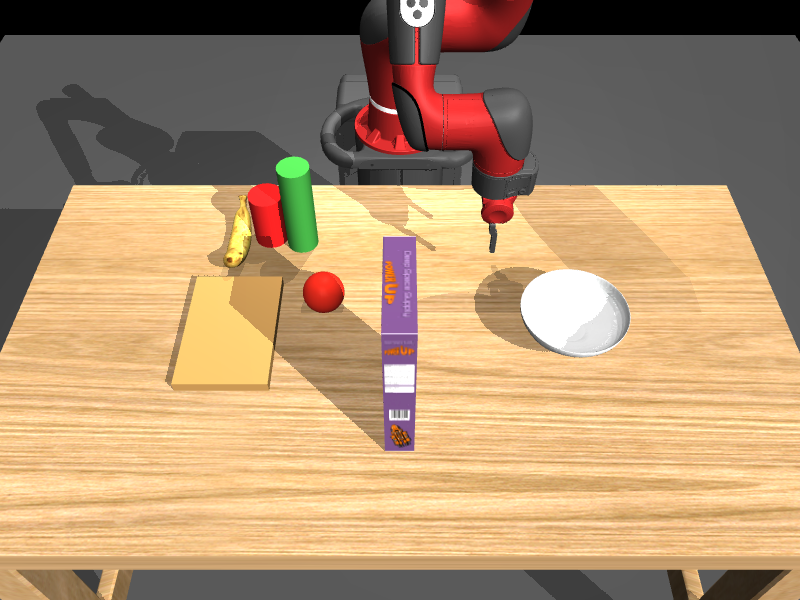} &
\sceneimg{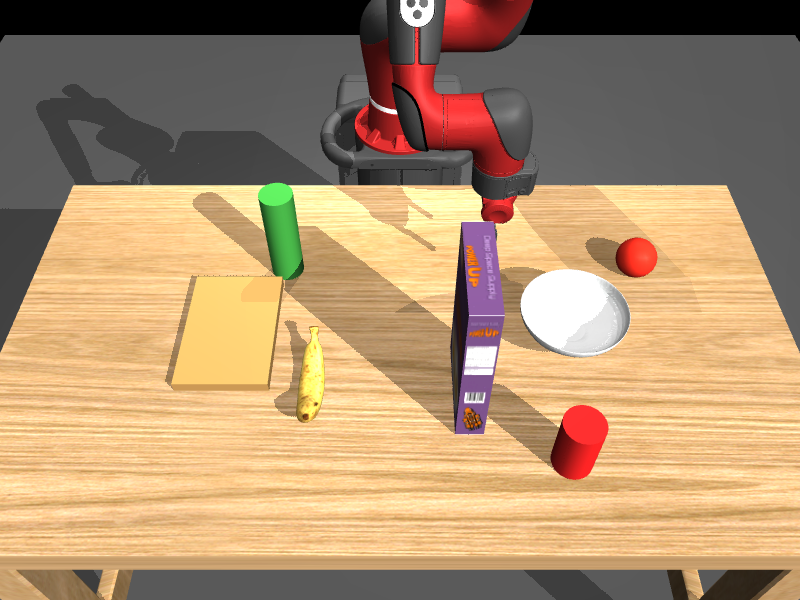} \\
\multicolumn{2}{c}{(b) Med-A: \emph{desk} ($M{=}3$)} &
\multicolumn{2}{c}{(e) Hard-A: \emph{meal\_assembly} ($M{=}5$)} \\[0.6em]
\sceneimg{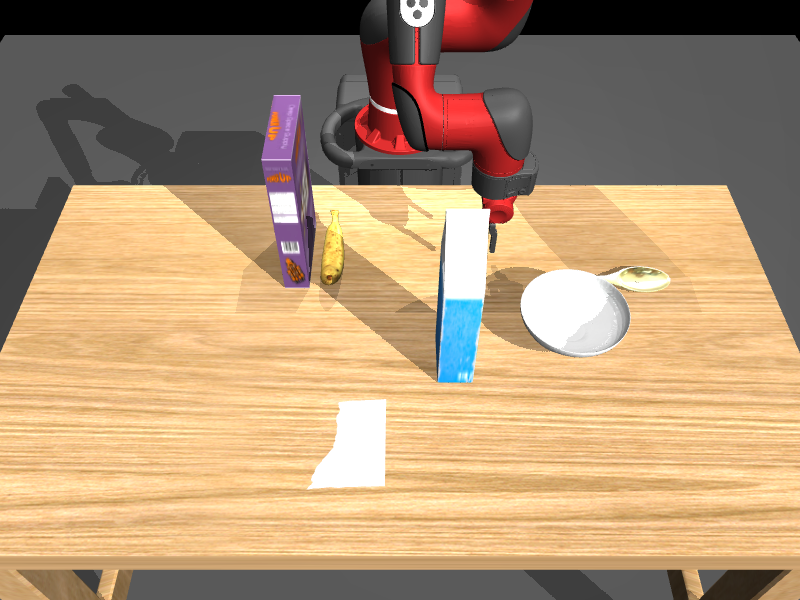} &
\sceneimg{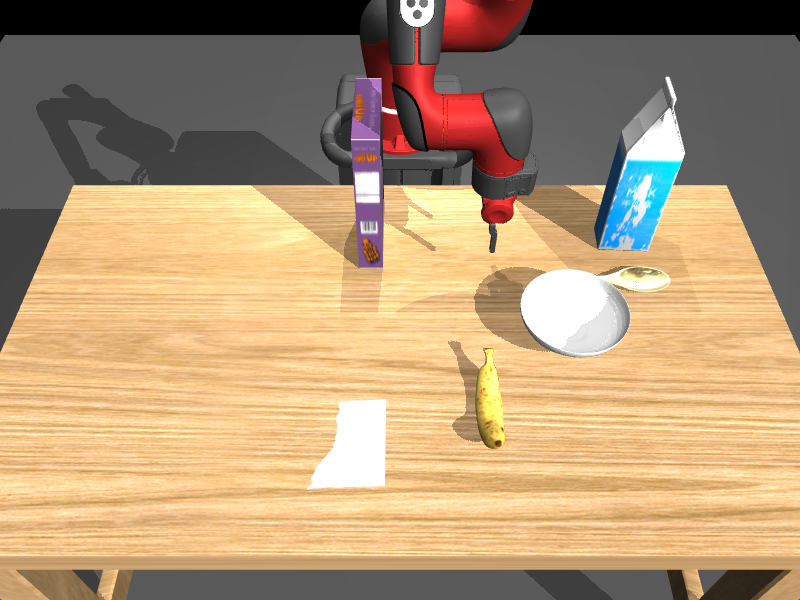} &
\sceneimg{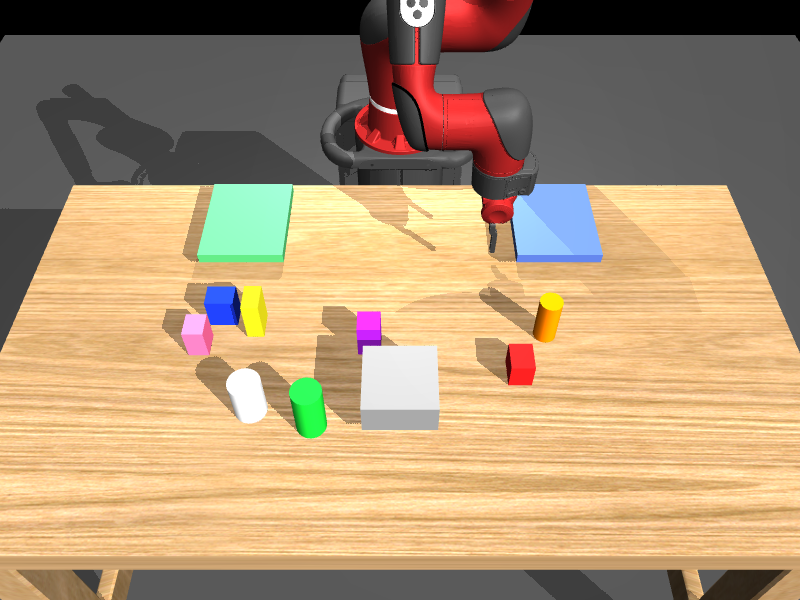} &
\sceneimg{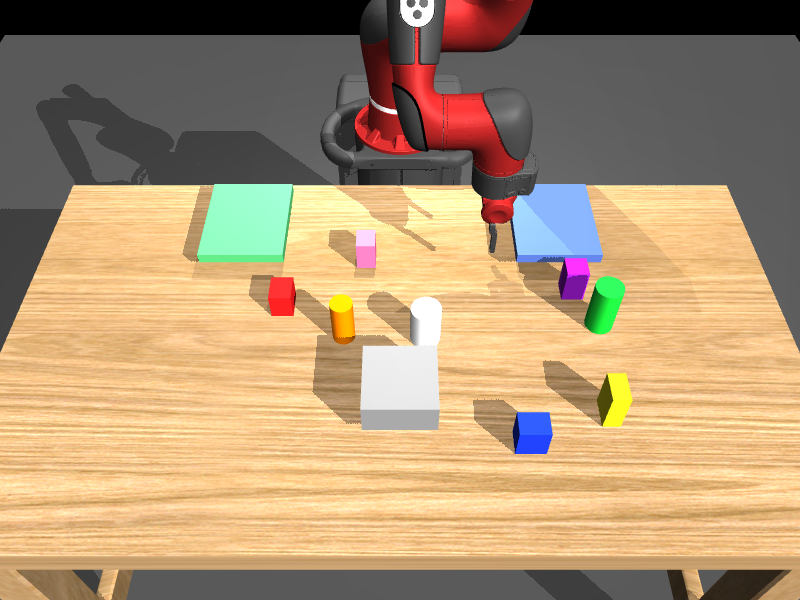} \\
\multicolumn{2}{c}{(c) Med-B: \emph{breakfast} ($M{=}3$)} &
\multicolumn{2}{c}{(f) Hard-B: \emph{cluttered} ($M{=}8$)} \\
\end{tabular}
\caption{\textbf{Random vs. SE(3)-optimized workspace layouts.} Each pair shows a random baseline (left) and an SE(3)-optimized layout (right) produced by MAP-Elites over $(x, y, \text{yaw})$ using the trajectory-margin objective. All layouts satisfy identical geometric constraints. By optimizing in SE(3), the method jointly places and orients objects to maintain separability of reachable grasp trajectories across $M$ candidate goals. Tasks share the same workspace and robot home pose.}
\label{fig:scenes}
\end{figure*}
\section{Experiments}
\label{sec:experiments}

We evaluate workspace optimization across six tabletop manipulation scenarios of increasing difficulty in simulation. The experiments are designed to address two research questions. \textbf{RQ 1:} Does workspace optimization improve goal inference compared to unoptimized (random) layouts? \textbf{RQ2} How does the approach scale with the number of candidate goals? We consider scenarios with $M{=}2$ to $M{=}8$ competing objects and report both inference performance and optimization cost.

\subsection{Experimental Setup}
\label{sec:setup}

\paragraph{Scenarios}
We construct six tabletop manipulation tasks using a shared Sawyer workspace (\cref{fig:scenes}), grouped into Easy, Medium, and Hard settings based on the number of candidate goals $M$. In each task, the robot must infer the human operator's intended object from early end-effector motion. All objects remain valid candidates at every decision point, requiring
the system to repeatedly disambiguate among $M$ alternatives. After each
pick, the selected object is returned to its initial position and the
inference process is reset, allowing multiple independent trials within
the same layout. To evaluate robustness across task sequences, we consider multiple permutations of object pick orderings. For tasks with $M \ge 5$, we
sample up to 120 permutations uniformly at random.


\paragraph{Grasp library}
Each object is assigned a fixed grasp pose in $\mathrm{SE}(3)$ from a hand-authored library that includes both top-down and side grasps. The choice of grasp type reflects the intended use of the object in downstream tasks (e.g., side grasps for objects that are typically lifted or poured, and top grasps for objects that are picked and placed from above). For each object, the grasp pose is parameterized relative to its geometry, with offsets computed from the object’s bounding box so that the gripper fingertips make contact with the object surface. This ensures consistent and physically feasible grasps across different object instances.

The grasp library is fixed across all experimental conditions; only object positions and orientations vary. Fixing grasps isolates the effect of workspace layout on goal inference and task execution, preventing confounding factors arising from grasp selection or manipulation strategy.

\paragraph{Conditions}
We compare two layout conditions that differ in how object positions are generated:
\begin{enumerate}
  \item \textbf{Random} ($n{=}30$ layouts per scene): Object positions $(x,y)$ are sampled uniformly within the workspace bounds, subject to minimum inter-object clearance ($12$\,cm), minimum distance from the robot end-effector ($15$\,cm radius), and table-edge margins. Object orientations are sampled uniformly from $[-\pi, \pi)$.
  \item \textbf{ME-optimized}: Layouts are generated using CMA-ME, producing a diverse archive of feasible solutions. We report results for the highest-performing elite in the archive. The behavior space is defined by the mean pairwise distance and centroid offset from the center of the workspace.
\end{enumerate}

\paragraph{Simulator}
All experiments are conducted using the MuJoCo simulator with the Sawyer robot controlled via 3D Jacobian pseudoinverse velocity control at $20$\,Hz. User input is simulated using the stochastic policy $\mathbf{u}_t = \boldsymbol{\pi}_h(\mathbf{x}_t, \mathbf{g}^*) + \boldsymbol{\epsilon}_t$ with Gaussian noise $\sigma_v {=} 0.03$\,m/s and a maximum end-effector speed of $0.05$\,m/s. The end-effector is driven toward the target grasp pose, and the inference loop terminates when either the posterior probability exceeds a commit threshold, the EE arrives at the goal pose, or a $30$\,s timeout is reached.

\paragraph{Observer}
Goal inference is performed using a Boltzmann-rational observer, which assigns probability to each candidate goal $g$ based on the efficiency of the observed trajectory: $P(g \mid \xi) \propto \exp(-\beta \cdot C(\xi, g))$ where $C$ is the path-efficiency cost ratio using an SE(3) distance metric with rotation weight $\lambda_R {=} 0.04$.  We use $\beta {=} 5$ and a default commit threshold $p_{\mathrm{thresh}} {=} 0.9$.

\paragraph{Feasibility check}
Before evaluation, each layout is valided for grasp feasibility. For each object, we test inverse kinematics (IK) reachability and collision-free execution at the corresponding grasp pose. Objects without a valid IK  solution or with collisions between the robot and non-target objects are marked as infeasible. During inference, infeasible grasps are excluded from candidate goals. However, they are counted as failures in the task success metric to reflect the practical impact of poor layout design.

\paragraph{Metrics}
\begin{itemize}
  \item \textbf{Argmax accuracy}: Fraction of picks where the final inferred goal matches the true goal.
  \item \textbf{Time to inference $T_{\mathrm{infer}}$}: Average time (in seconds) from end-effector motion onset to the first commitment over all pick steps.
\end{itemize}

\section{Results}
\label{sec:results}

\subsection{Inference Accuracy and Speed}
\label{sec:rq1_results}

We evaluate whether workspace optimization improves intent inference (RQ1) and how performance scales with the number of candidate goals (RQ2). \cref{tab:main_results} summarizes results across all six scenarios.

\begin{table}[t]
\centering
\caption{Trajectory-margin slack, argmax accuracy, and time to inference across tiers. Random denotes $30$ random feasible layouts (mean $\pm$ std). ME-optimized is the best layout found with MAP-Elites. Slack is the worst-case pairwise trajectory-margin slack computed at Bonferroni-corrected significance $\alpha{=}0.05$; positive slack implies a nominal argmax-accuracy guarantee of $\geq 1 - \alpha = 95\%$ under the straight-line, linear-Gaussian approximation (\cref{eq:traj_slack}).}
\label{tab:main_results}
\small
\setlength{\tabcolsep}{5pt}
\begin{tabular}{@{}llccccc@{}}
\toprule
\textbf{Tier} & \textbf{Layout} & $M$ &
  Slack &
  $T_{\mathrm{infer}}$ (s) &
  Argmax acc. \\
\midrule
\multirow{3}{*}{Easy}
  & Random    & 2 & --- & $6.1 \pm 2.4$ & $92\%$ \\
  & ME-optimized   &   & $8.03$ & $\mathbf{5.4}$ & $\mathbf{100\%}$ \\
\midrule
\multirow{3}{*}{Med-A}
  & Random    & 3 & ---  & $\mathbf{6.9 \pm 3.4}$ & $91\%$ \\
  & ME-optimized   &   & $4.68$ & $9.4$ & $\mathbf{100\%}$ \\
\midrule
\multirow{3}{*}{Med-B}
  & Random    & 3 & --- & $9.9 \pm 2.8$ & $\mathbf{100\%}$ \\
  & ME-optimized   &   & 5.14 & $\mathbf{9.5}$          & $\mathbf{100\%}$ \\
\midrule
\multirow{3}{*}{Med-C}
  & Random    & 3 & ---& $7.7 \pm 3.9$ & $93\%$ \\
  & ME-optimized   &   & 6.07 & $\mathbf{5.1}$ & $\mathbf{100\%}$ \\
\midrule
\multirow{3}{*}{Hard-A}
  & Random    & 5 & --- & $12.5 \pm 3.0$ & $88\%$ \\
  & ME-optimized &   & 1.74 & $\mathbf{10.6}$ & $\mathbf{100\%}$ \\
\midrule
\multirow{3}{*}{Hard-B}
  & Random    & 8 & --- & $10.9 \pm 3.6$ & $25\%$ \\
  & ME-optimized & & -0.99 & $\mathbf{9.6}$ & $\mathbf{54\%}$ \\
\bottomrule
\end{tabular}
\end{table}

\paragraph{Accuracy}
ME-optimized layouts consistently achieve higher argmax accuracy than random baselines across all scenarios, directly supporting \textbf{RQ1}. While random layouts degrade from $88$\% (Easy, $M{=}2$) to $67$\% (Hard-B, $M{=}8$), optimized layouts maintain near-perfect performance ($100$\% in most cases and $97$\% in Hard-B). 

This gap widens as the number of candidate goals increases, providing strong evidence for \textbf{RQ2}. As the workspace becomes more cluttered and the number of competing hypotheses grows, random placements increasingly produce ambiguous trajectories, whereas optimized layouts preserve distinguishability between goals. This demonstrates that workspace optimization is particularly beneficial in high-ambiguity regimes.

\paragraph{Time to inference}
Workspace optimization reduces the time required for intent inference in moderate-difficulty scenarios, further supporting \textbf{RQ1}. In Med-A and Med-C, optimized layouts enable earlier disambiguation (e.g., $4.1$s vs. $7.2$s and $3.5$s vs. $7.3$s), indicating that increased goal separation allows the robot to reach confident predictions with less trajectory evidence. This trend is less consistent in Easy, where improvements in accuracy do not always translate to faster inference. This suggests that while margin increases improve correctness, they do not always accelerate confidence accumulation.

As the number of candidate goals increases (RQ2), the time-to-inference gap narrows. In Hard-A and Hard-B, both random and optimized layouts require longer trajectories to disambiguate among many competing goals. However, optimized layouts maintain substantially higher accuracy, indicating that in high-$M$ regimes, the primary benefit shifts from speed to robustness.
\begin{figure*}[t]
    \centering
    \begin{minipage}[b]{0.05\textwidth}
        \vfill
        \centering
        \raisebox{0.5\height}{\rotatebox{90}{\textbf{BASELINE}}}
        \vfill
    \end{minipage}%
    \begin{minipage}[b]{0.93\textwidth}
        \centering
        \begin{subfigure}[b]{0.32\textwidth}
            \centering
            \includegraphics[width=\linewidth]{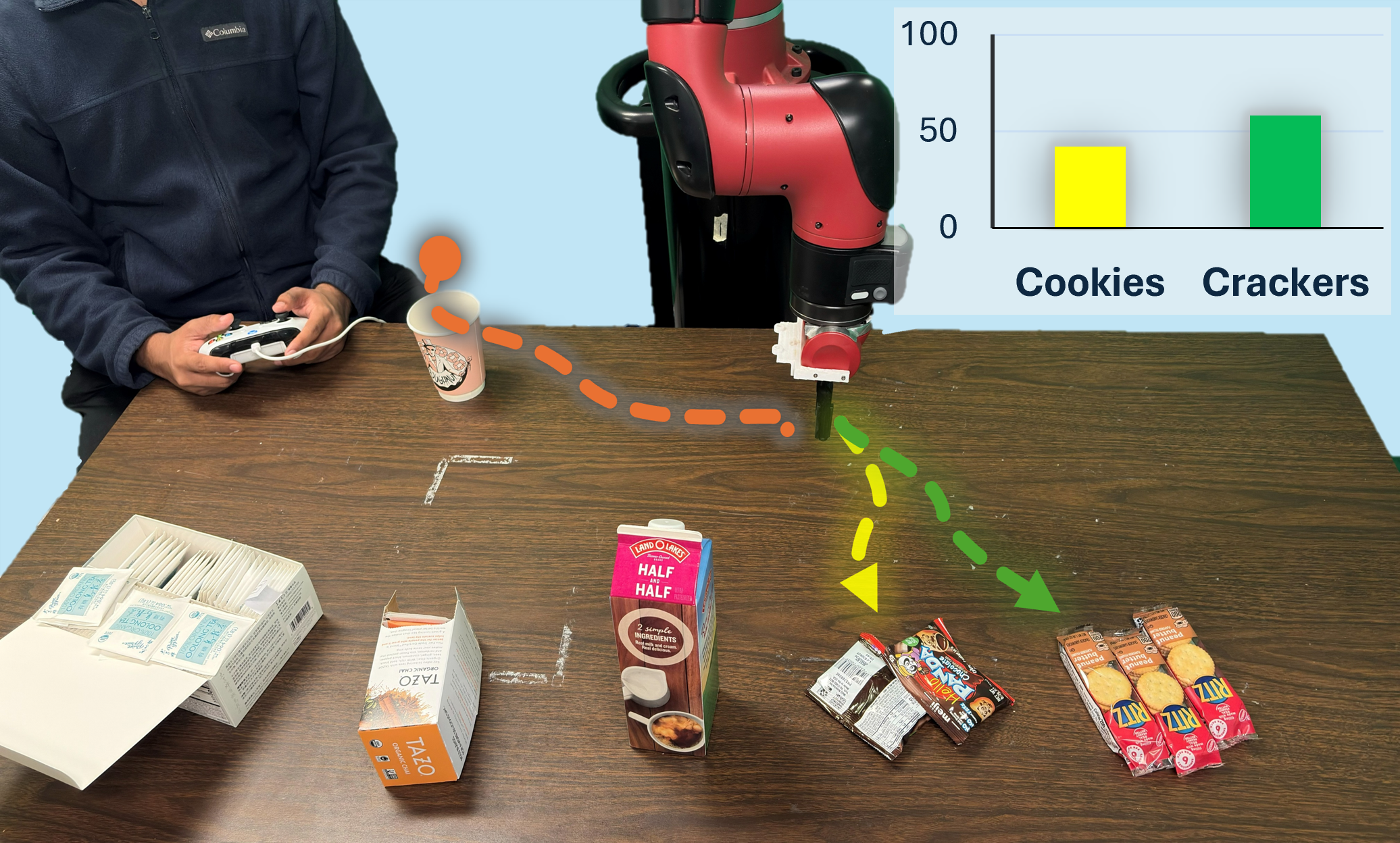}
        \end{subfigure}\hfill
        \begin{subfigure}[b]{0.32\textwidth}
            \centering
            \includegraphics[width=\linewidth]{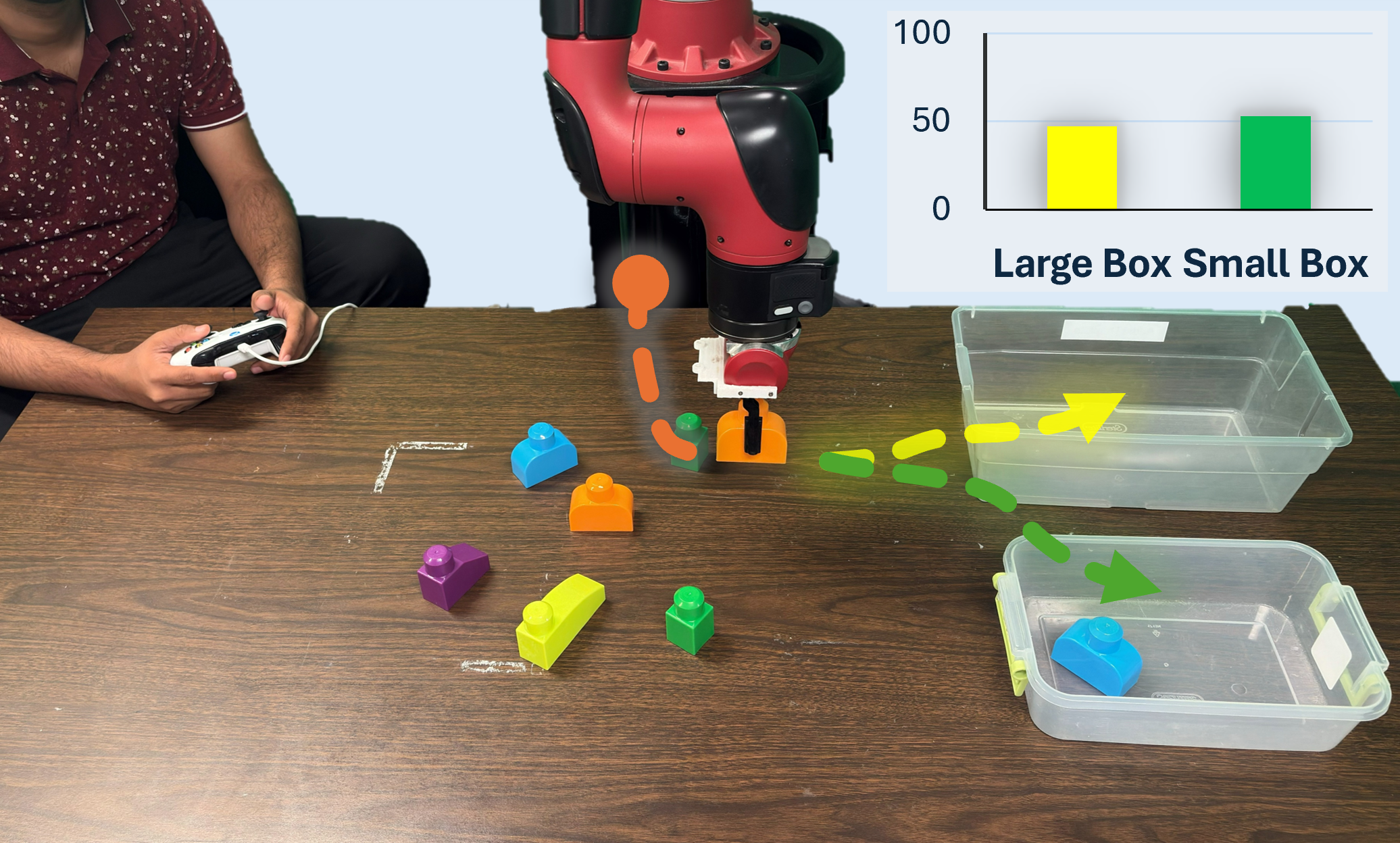}
        \end{subfigure}\hfill
        \begin{subfigure}[b]{0.32\textwidth}
            \centering
            \includegraphics[width=\linewidth]{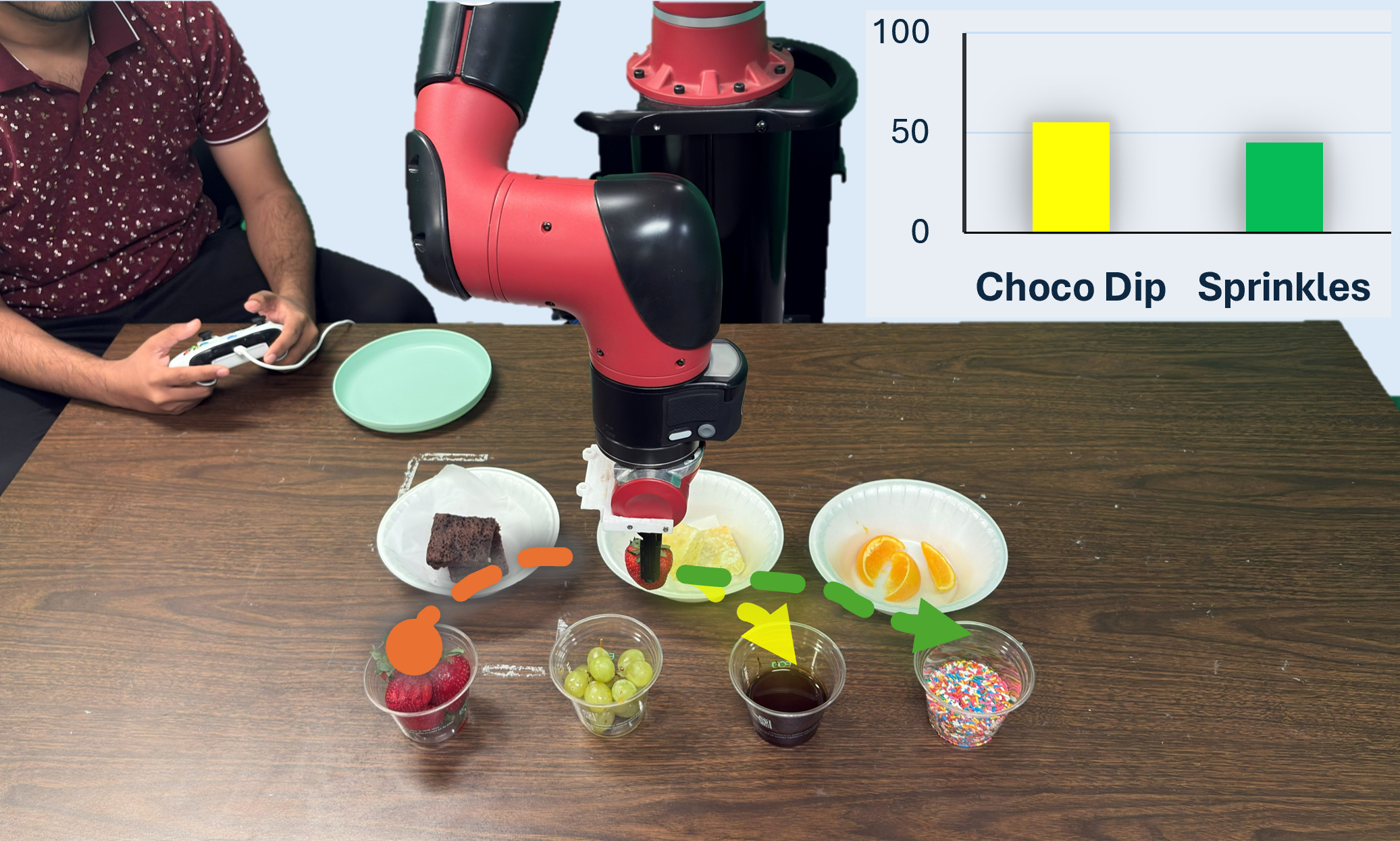}
        \end{subfigure}
    \end{minipage}

    \vspace{1em}

    \begin{minipage}[b]{0.05\textwidth}
        \vfill
        \centering
        \raisebox{0.03\height}{\rotatebox{90}{\textbf{ME-OPTIMIZED}}}
        \vfill
    \end{minipage}%
    \begin{minipage}[b]{0.93\textwidth}
        \centering
        \begin{subfigure}[b]{0.32\textwidth}
            \centering
            \includegraphics[width=\linewidth]{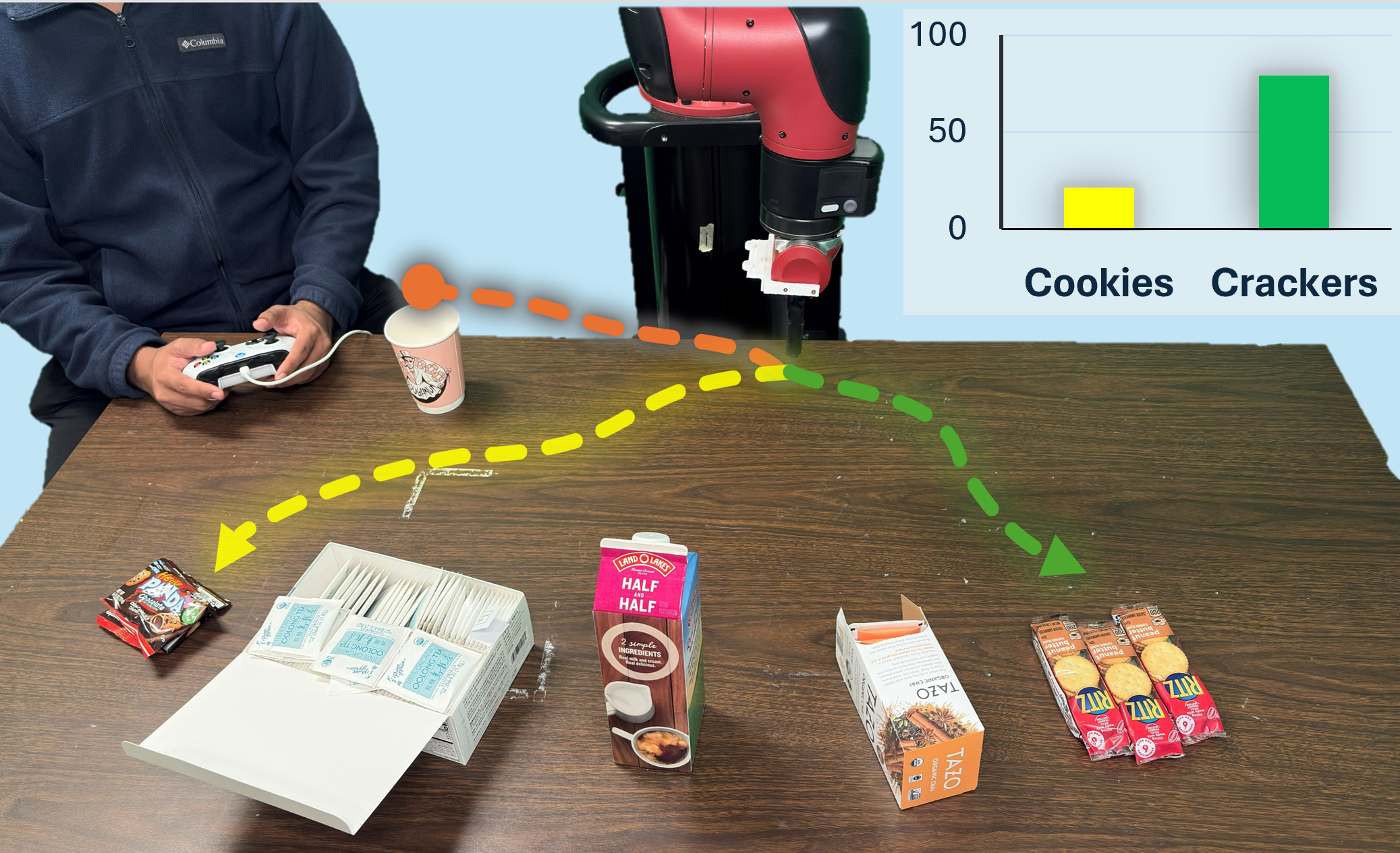}
        \end{subfigure}\hfill
        \begin{subfigure}[b]{0.32\textwidth}
            \centering
            \includegraphics[width=\linewidth]{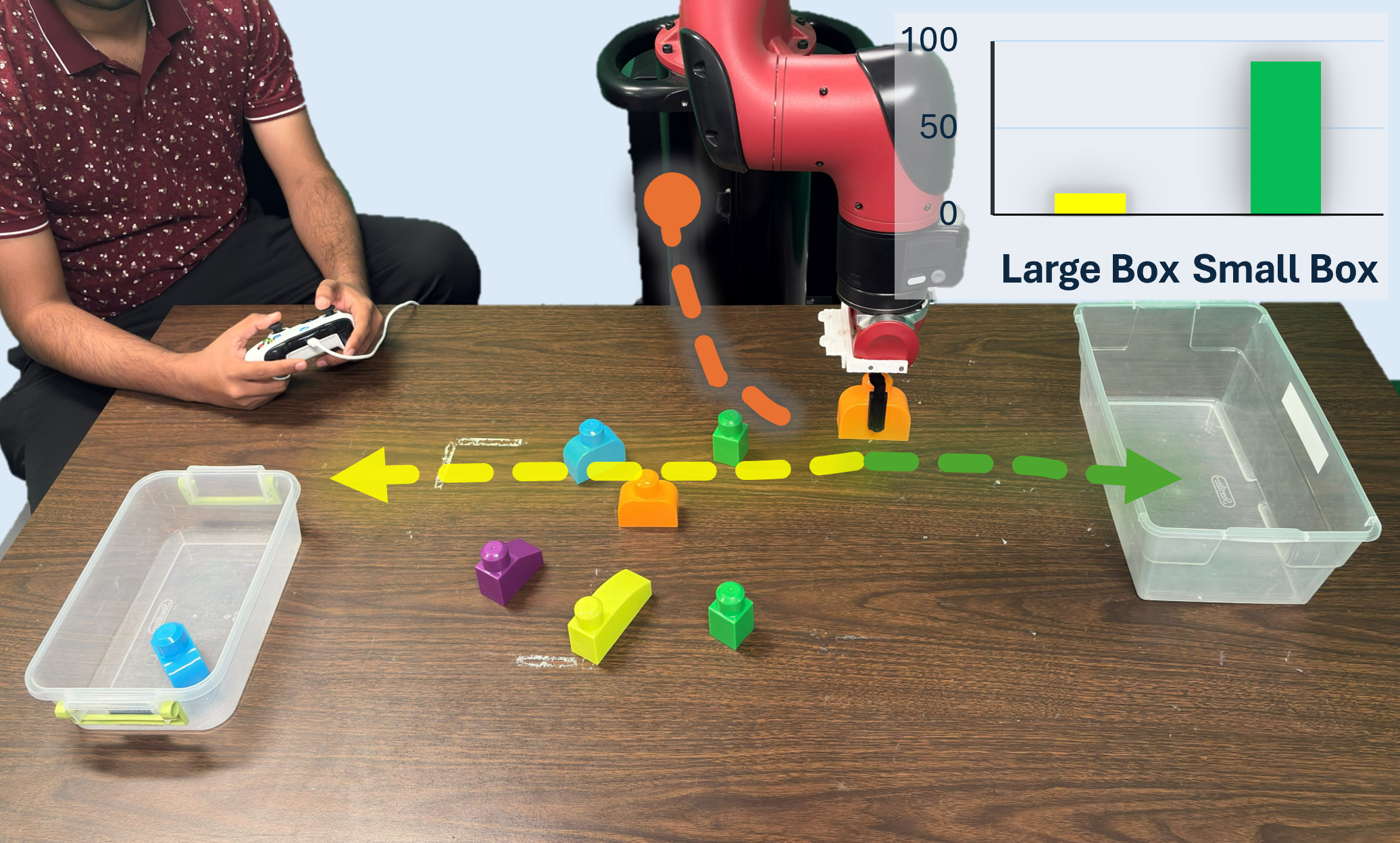}
        \end{subfigure}\hfill
        \begin{subfigure}[b]{0.32\textwidth}
            \centering
            \includegraphics[width=\linewidth]{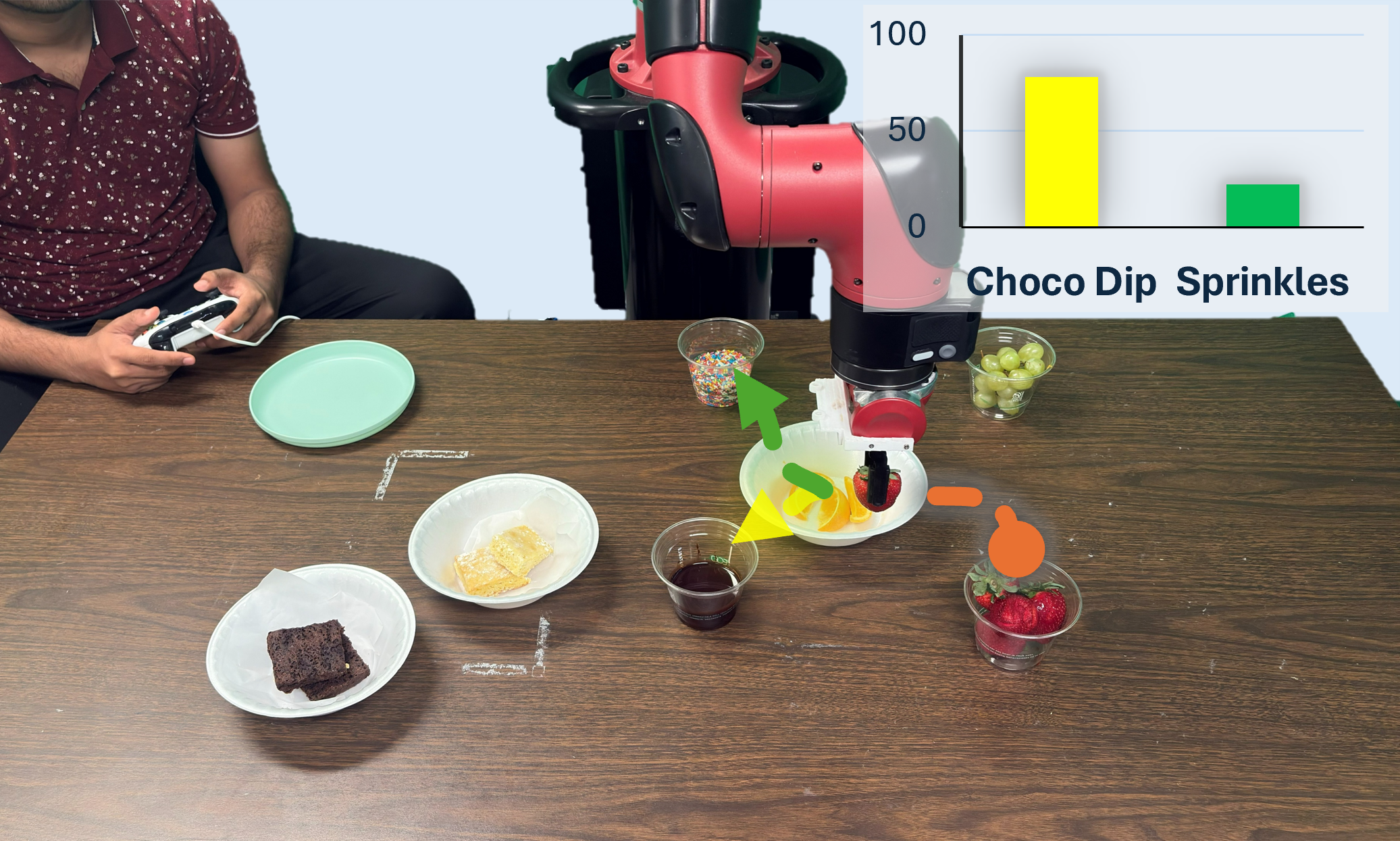}
        \end{subfigure}
    \end{minipage}
    \caption{Illustration of three shared autonomy scenarios: (a) grabbing snacks for tea, (b) sorting LEGO blocks into containers, and (c) reaching the dips for a strawberry. The first row (BASELINE) presents intuitive arrangements, where similar or related objects are grouped together. The second row (ME-OPTIMIZED) presents the configurations produced by the MAP-Elites algorithm, which optimizes the workspace for improved intent prediction.}
    \label{fig:sawyer}
\end{figure*}

\subsection{Optimizer Compute Cost}
\label{sec:rq2_results}
We analyze how the optimization scales with the number of candidate goals (RQ2). \cref{tab:compute} reports wall-clock time for MAP-Elites across all tiers.

Optimization time ranges from $340$--$920$\,s per scene and does not increase monotonically with $M$. Instead, runtime is influenced by the structure of the search space, while archive coverage remains consistently high ($93$--$100$\%). This indicates that the approach scales favorably in practice, even as the number of candidate goals increases.

A key advantage of MAP-Elites is the diversity of solutions produced. With hundreds of feasible layouts spanning the behavior space, practitioners can select among top-performing candidates at deployment time. This provides robustness to potential mismatch between the optimization objective and the runtime inference model.

\begin{table}[t]
\centering
\caption{Optimizer compute cost (wall-clock seconds, single seed). We also report the number of solutions in the MAP-Elites archive and the archive coverage.}
\label{tab:compute}
\small
\begin{tabular}{@{}lcccc@{}}
\toprule
\textbf{Tier} & $M$ & ME (s) & Elites & Coverage \\
\midrule
Easy   & 2 & 597 & 374 & 93.5\% \\
Med-A  & 3 & 920 & 399 & 99.8\% \\
Med-B  & 3 & 653 & 400 & 100\%  \\
Med-C  & 3 & 727 & 397 & 99.3\% \\
Hard-A & 5 & 461 & 400 & 100\%  \\
Hard-B & 8 & 340 & 400 & 100\%  \\
\bottomrule
\end{tabular}
\end{table}

\subsection{Real-World System Demonstration}

We demonstrate our framework in three scenarios: (a) making tea with snacks, (b) sorting LEGO blocks, and (c) assistive feeding with brownies and fruits (\cref{fig:sawyer}). In the tea-making task, the user prepares the tea and then selects one of several snacks. In the LEGO sorting task, the robot assists the user in sorting colored LEGO blocks, without prior knowledge of which container corresponds to each color. In the assistive feeding task, the user chooses between fruits (e.g., strawberry or grape) and brownies, with the option to dip them into chocolate or sprinkles before feeding.

In the baseline condition, objects are placed in an intuitive manner, where similar or related items are grouped together (e.g., teas placed side-by-side, or fruits clustered together). While this arrangement appears natural, it can make intent inference more ambiguous for the robot--causing lower confidence during teleoperation and delaying assistance.

In contrast, the optimized condition applies our workspace optimization algorithm to generate object configurations that improve goal separability. Unlike our simulation experiments, which assume a fixed set of candidate goals, real-world tasks are sequential and context-dependent. We exploit this structure by optimizing only the subset of relevant goals at each stage of the task, as defined by the task graph. This produces layouts that adapt to the current decision context, enabling more reliable disambiguation as the task progresses. By strategically placing and orienting objects to maximize predictability, the robot infers the user’s intent earlier and transitions more effectively to autonomous assistance. These demonstrations illustrate how environment design can be leveraged to improve shared autonomy in realistic, multi-stage tasks, and highlight the potential of our approach for broader human-robot collaboration settings.
\section{Conclusion}
We presented a framework for improving intent inference in shared autonomy through workspace design. Rather than treating the environment as fixed, we showed that object placement fundamentally shapes the difficulty of the inference problem. By formulating workspace design as an optimization problem over goal separability, we derived a margin-based probabilistic correctness condition that guarantees the true goal is identified with probability at least $1 - \alpha$ under bounded noise.

Our results demonstrate that optimized workspaces consistently improve inference accuracy across a range of scenarios and maintain high performance as the number of candidate goals increases. While gains in time to inference are most pronounced in moderate-difficulty settings, the primary benefit in more complex scenarios is improved robustness, where optimized layouts preserve reliable goal disambiguation despite increased ambiguity. These findings highlight a key advantage of environment design: it enables principled control over inference difficulty without modifying the underlying inference algorithm.

We further validated our approach in real-world tabletop tasks, showing that task-aware workspace optimization can adapt to multi-stage interactions by focusing on relevant subsets of goals. This demonstrates the practicality of our framework and its applicability to realistic human-robot collaboration settings. 

Overall, this work establishes environment design as a complementary axis for improving shared autonomy systems. By explicitly shaping the geometry of the inference problem, we can achieve more reliable, predictable, and user-aligned robot behavior. We plan to conduct user studies to evaluate the impact of optimized workspaces on usability, trust, and task efficiency as future work.
%
%
%
%
%
\appendix
\subsection{Local validity of the linearized margin}
\label{app:linearization_validity}

The pairwise cost difference $\delta c(\mathbf{g})$ is, in general, a nonlinear function of the trajectory and terminal state. As a result, the log-posterior margin is not exactly Gaussian under Gaussian execution noise. To obtain a tractable probabilistic characterization, we approximate $\delta c$ using a first-order Taylor expansion around the nominal terminal state $\bar{\mathbf{x}}_T$. We now show that this linearization is locally accurate with high probability, provided the execution noise remains within a neighborhood where the curvature of $\delta c(\mathbf{g})$ is bounded.

\paragraph{Proposition.}
Let $f_g(\mathbf{x}_T) := \delta c(\mathbf{g}; \mathbf{x}_T)$ denote the pairwise cost difference as a function of the terminal state. Assume $f_g$ is twice continuously differentiable in a neighborhood $\mathcal{N}_g$ of $\bar{\mathbf{x}}_T$, and that its Hessian is bounded:
\begin{equation}
  \|\nabla^2 f_g(\mathbf{x}_T)\|_2 \le L_g,
  \qquad \forall \mathbf{x}_T \in \mathcal{N}_g.
\end{equation}
Then for any $\mathbf{x}_T \in \mathcal{N}_g$, the second-order Taylor expansion gives
\begin{equation}
  f_g(\mathbf{x}_T)
  =
  f_g(\bar{\mathbf{x}}_T)
  +
  \nabla f_g(\bar{\mathbf{x}}_T)^\top (\mathbf{x}_T - \bar{\mathbf{x}}_T)
  +
  R_g(\mathbf{x}_T),
\end{equation}
where the remainder satisfies
\begin{equation}
  |R_g(\mathbf{x}_T)|
  \le
  \frac{L_g}{2}
  \|\mathbf{x}_T - \bar{\mathbf{x}}_T\|^2.
\end{equation}

Suppose $\mathbf{x}_T \sim \mathcal{N}(\bar{\mathbf{x}}_T, \Sigma_{x_T})$,
and let $r_\alpha > 0$ satisfy
\begin{equation}
  \Pr\!\left(\|\mathbf{x}_T - \bar{\mathbf{x}}_T\| \le r_\alpha\right)
  \ge 1 - \alpha.
\end{equation}
If the ball $\{\mathbf{x}_T : \|\mathbf{x}_T - \bar{\mathbf{x}}_T\| \le r_\alpha\}$ is contained in $\mathcal{N}_g$, then with probability at least $1-\alpha$,
\begin{equation}
  |R_g(\mathbf{x}_T)|
  \le
  \frac{L_g}{2} r_\alpha^2.
\end{equation}

\paragraph{Implication.}
Within this high-probability region, the nonlinear cost difference $f_g(\mathbf{x}_T)$ is well-approximated by its linearization. Therefore,
the log-posterior margin can be approximated as a Gaussian random variable with bounded error. In particular, if
\begin{equation}
  \frac{L_g}{2} r_\alpha^2 \ll m_g,
\end{equation}
where $m_g$ is the nominal separation margin, then the linearized model preserves the sign and magnitude of the margin with high probability. This justifies using the Gaussian approximation to derive a sufficient chance constraint for correct goal inference.

\subsection{Quality-Diversty Optimization Details} \label{app:qd}

This section provides implementation details for the quality-diversity (QD) optimization procedure described in Section~\ref{sec:qd}.

\paragraph{Solution parameterization.}
Each candidate layout is represented by the $(x,y, yaw)$ positions of the $M$ movable objects, yielding a $3M$-dimensional decision vector.

\paragraph{Algorithm.}
We use CMA-ME~\cite{fontaine2020covariance}, a variant of MAP-Elites that employs Covariance Matrix Adaptation Evolution Strategy (CMA-ES) emitters to explore the solution space. We use three emitters operating in parallel, each with a batch size of 30 and initial step size $\sigma_0 = 0.08$. Emitters are ranked using the two-stage improvement criterion described in~\cite{fontaine2020covariance}.

\paragraph{Behavior space discretization.}
The archive is a $20 \times 20$ grid defined over two behavioral dimensions:
\begin{itemize}
  \item mean pairwise distance between objects, and
  \item centroid offset from the workspace center.
\end{itemize}
The range of each dimension is determined by sampling $10{,}000$ random layouts and computing the 5th and 95th percentiles of the resulting feature values.

\paragraph{Constraint handling.}
Candidate layouts are required to satisfy the following constraints:
\begin{itemize}
  \item \textbf{Workspace bounds:} 

  $x \in [0.30, 0.85]~\mathrm{m}, \quad
  y \in [-0.45, 0.45]~\mathrm{m}.$
  \item \textbf{Minimum inter-object clearance:}
  
  $\|\mathbf{g}_i - \mathbf{g}_j\| \ge 0.12~\mathrm{m} \quad \forall i \neq j.$
  \item \textbf{Minimum distance from robot end-effector:} 
  
  $\|\mathbf{g}_i - \mathbf{x}_0\| \ge 0.15~\mathrm{m}.$
\end{itemize}
Constraint violations are handled via additive penalties in the objective function, discouraging infeasible layouts from entering the archive.

\paragraph{Evaluation.}
Each candidate layout is evaluated using the trajectory margin slack objective defined in Section~\ref{sec:trajectory_margin}. The archive stores the layout with the highest objective value in each cell.

\bibliographystyle{IEEEtran}
\bibliography{main}

\end{document}